\documentclass[journal]{IEEEtran}
\IEEEoverridecommandlockouts

\pdfoutput=1
\usepackage{cite}
\usepackage{amsmath,amssymb,amsfonts}
\usepackage{algorithmic}
\usepackage{graphicx}
\usepackage{textcomp}
\usepackage{xcolor}
\usepackage{siunitx}
\usepackage{booktabs}
\usepackage{tabularx}
\usepackage{multirow}
\usepackage{array}
\usepackage{makecell}
\usepackage[font=footnotesize]{caption}
\usepackage{soul}
\usepackage[caption=false,font=footnotesize,labelfont=rm,textfont=rm]{subfig}

\usepackage{tikz,xcolor}
\usepackage[hidelinks]{hyperref}

\definecolor{lime}{HTML}{A6CE39}
\DeclareRobustCommand{\orcidicon}{%
    \begin{tikzpicture}
    \draw[lime, fill=lime] (0,0) 
    circle [radius=0.16] 
    node[white] {{\fontfamily{qag}\selectfont \tiny ID}};    \draw[white, fill=white] (-0.0625,0.095) 
    circle [radius=0.007];    \end{tikzpicture}
    \hspace{-2mm}}
\foreach \x in {A, ..., Z}{%
    \expandafter\xdef\csname orcid\x\endcsname{\noexpand\href{https://orcid.org/\csname orcidauthor\x\endcsname}{\noexpand\orcidicon}}
    }

\hypersetup{
    colorlinks=false,
    linkcolor=black, 
    urlcolor=black 
    citecolor=black, 
    filecolor=black, 
    menucolor=black 
}

\def\BibTeX{{\rm B\kern-.05em{\sc i\kern-.025em b}\kern-.08em
    T\kern-.1667em\lower.7ex\hbox{E}\kern-.125emX}}
    
\begin{document}

\title{Adaptive Integral Sliding Mode Control for Attitude Tracking of Underwater Robots With Large Range Pitch Variations in Confined Space\\
\thanks{Xiaorui Wang and Zeyu Sha are with the Department of Advanced Manufacturing and Robotics, College of Engineering, Peking University, Beijing, 100871, China (email: \href{mailto:jnswxr@stu.pku.edu.cn}{jnswxr@stu.pku.edu.cn} and \href{mailto:schahzy@stu.pku.edu.cn}{schahzy@stu.pku.edu.cn})

Feitian Zhang is with the Department of Advanced Manufacturing and Robotics, and the State Key Laboratory of Turbulence and Complex Systems, College of Engineering, Peking University, Beijing, 100871, China (email: \href{mailto:feitian@pku.edu.cn}{feitian@pku.edu.cn})}
}

\author{Xiaorui Wang, Zeyu Sha, and Feitian Zhang\textsuperscript{*}}

\maketitle
\pagestyle{empty}  
\thispagestyle{empty} 

\begin{abstract}
Underwater robots play a crucial role in exploring aquatic environments. The ability to flexibly adjust their attitudes is essential for underwater robots to effectively accomplish tasks in confined space. However, the highly coupled six degrees of freedom dynamics resulting from attitude changes and the complex turbulence within limited spatial areas present significant challenges. To address the problem of attitude control of underwater robots, this letter investigates large-range pitch angle tracking during station holding as well as simultaneous roll and yaw angle control to enable versatile attitude adjustments. Based on dynamic modeling, this letter proposes an adaptive integral sliding mode controller (AISMC) that integrates an integral module into traditional sliding mode control (SMC) and adaptively adjusts the switching gain for improved tracking accuracy, reduced chattering, and enhanced robustness. The stability of the closed-loop control system is established through Lyapunov analysis. Extensive experiments and comparison studies are conducted using a commercial remotely operated vehicle (ROV), the results of which demonstrate that AISMC achieves satisfactory performance in attitude tracking control in confined space with unknown disturbances, significantly outperforming both PID and SMC.
\end{abstract}


\section{Introduction}

\begin{figure*}[tbp]
\centering
\includegraphics[width=0.97\linewidth]{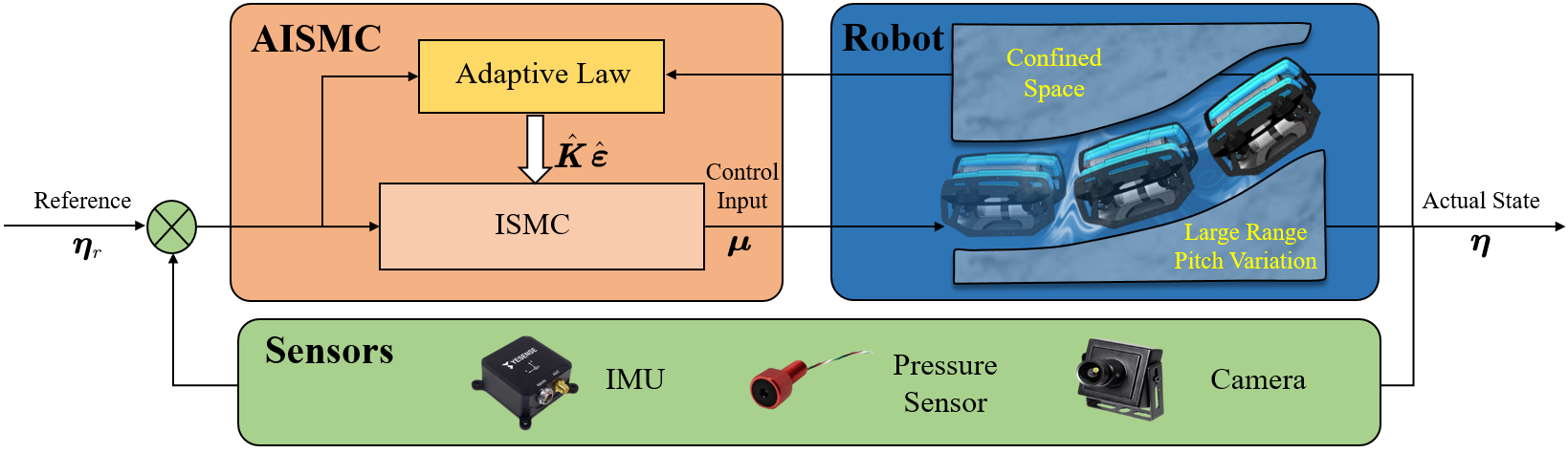}
\caption{AISMC attitude closed-loop control framework for underwater robots, enabling a wide range of pitch variations in confined space.}
\label{fig1}
\end{figure*}

\IEEEPARstart{T}{he} Earth's surface is predominantly covered by water, accounting for 71\% of its total area. These vast bodies of water remain largely unexplored and hold immense potential as reservoirs of resources. Human exploration in this domain is incessant, with underwater robots increasingly employed to undertake diverse tasks. Attitude control plays a pivotal role in ensuring the successful completion of these tasks, especially for underwater robots holding their stations and operating in complex environments, necessitating numerous researches efforts\cite{NMPC2023}, \cite{ISMC-ECC2021}, \cite{SMC-RAL}, \cite{RAL-vision}, \cite{shark-AUV}, \cite{MPC-RAL2021}, \cite{RISE-TRO-2014}, \cite{RAL-robust}, \cite{PID-RAL}. For instance, Walker \textit{et al.} investigated orientation control of underwater robots when holding position amidst wave-induced oceanic disturbances \cite{NMPC2023}, while Benzon explored heading tracking of an underwater robot removing moss in open waters \cite{ISMC-ECC2021}. 

The existing designs for the attitude control of underwater robots are diverse, including but not limited to proportional-integral-derivative (PID) control \cite{PID-RAL}, \cite{PID2019}, model predictive control (MPC) \cite{NMPC2023}, \cite{MPC-RAL2021}, \cite{Obstacle-MPC}, adaptive control \cite{AC2019}, \cite{GSTA-ESO2020}, and sliding mode control (SMC) \cite{SMC-RAL},\cite{OpenAUV2022}, \cite{SMC2015}, \cite{SMC2023}, \cite{SMC2011}. Among them, SMC stands out for its robustness and disturbance rejection capabilities. However, traditional SMC has its limitations. The high uncertainty in robot dynamics and the presence of complex fluid perturbations often result in reduced tracking accuracy and increased steady-state errors. Researchers such as Benzon \cite{ISMC-ECC2021}, Tugal \cite{ISMC-ICRA2022}, and Raygosa-Barahon \cite{ISMC2011} explored the use of integral sliding mode control (ISMC) to improve tracking accuracy in the presence of complex flow disturbances. Additionally, SMC often requires an advance estimation of the upper bound of system disturbance to ensure stability\cite{ISMC-ECC2021}, \cite{OpenAUV2022}. However, setting an excessively high upper bound generally compromises control accuracy and potentially causes severe chattering issues. Moreover, both system and environmental disturbances are often time-varying. To address these challenges, researchers like Fei
\cite{ASMC2019_1}, Raja \cite{ASMC2019_2}, and Plestan \cite{ASMC2013} developed adaptive sliding mode control (ASMC) that adjusted the switching gain of SMC based on changes in the system states and estimated external disturbances.

While most existing studies on attitude control of station-holding underwater robots primarily focused on yaw control, assuming fixed pitch and roll angles, the ability to adapt their pitch over a large range of angles is crucial for underwater robots to effectively complete tasks in confined space environments, such as navigating through a narrow tunnel as depicted in Fig.~\ref{fig1}. 
The attitude control of station-holding underwater robots with a large-range pitch variation is non-trivial due to several factors, including the strong coupling and high nonlinearity of robot dynamics \cite{ISMC-ICRA2022}, inaccuracies in hydrodynamic modeling \cite{AC2019}, and the presence of unstable equilibrium corresponding to large pitch angles \cite{pitch_reduction}. Furthermore, when operating within confined space, the propeller of the underwater robot generates complex wake flows mixed with water rebounding from walls nearby. This phenomenon leads to the formation of complex turbulence,  posing a significant challenge to the attitude control of underwater robots \cite{MPC-RAL2021}.

This letter proposes the adaptive integral sliding mode control (AISMC) for the attitude tracking control of underwater robots with large-range pitch variations in confined space, which takes advantages of both ISMC and ASMC. Figure~\ref{fig1} provides an overview of the proposed AISMC control framework. AISMC comprehensively considers control accuracy, robustness, disturbance rejection capability, and real-time estimation of system uncertainties, enabling the robot to effectively follow desired attitude trajectories in confined and dynamic environments. Taking BlueROV2 as an example underwater robot, extensive experiments are conducted to validate the proposed control design against selected benchmark controllers for comparison. 

The contribution of this letter is twofold. First, unlike other studies primarily focusing on the yaw tracking control of station-holding underwater robots with fixed pitch and roll angles, this letter addresses a more challenging task of attitude control of underwater robots with large-range pitch variations in confined space. To tackle such a demanding control design task and effectively handle complex and unknown disturbances, a novel AISMC control scheme is proposed for underwater robots to ensure both robustness and tracking accuracy. Second, the stability of the control system is rigorously established using Lyapunov analysis, and the effectiveness of the proposed attitude tracking control is validated through extensive experiments and comparison studies.

\section{Experimental Platform and Setup}\label{section2}

\subsection{BlueROV2}
This letter uses the BlueROV2 Heavy, an off-the-shelf underwater robot, as the experimental robot platform. The robot, employing a cable connection to an onshore computer for remote operation and data transmission, is equipped with an onboard inertial measurement unit and a pressure sensor to measure its attitude and depth, respectively. Table~\ref{tab1} provides detailed specifications of the sensors used. With eight independent Blue Robotics T200 thrusters (four horizontally oriented and four vertically oriented), the robot has complete control over its six degrees of freedom motions. The BlueROV2 Heavy incorporates a Raspberry PI 3 as an intermediate hub between the upper computer and the lower Pixhawk module. The MAVlink protocol is utilized to transmit command instructions to the lower Pixhawk, while simultaneously updating real-time status back to the upper computer.

\subsection{Experimental Setup}
\begin{figure}[tbp]
\centering
\includegraphics[width=0.95\linewidth]{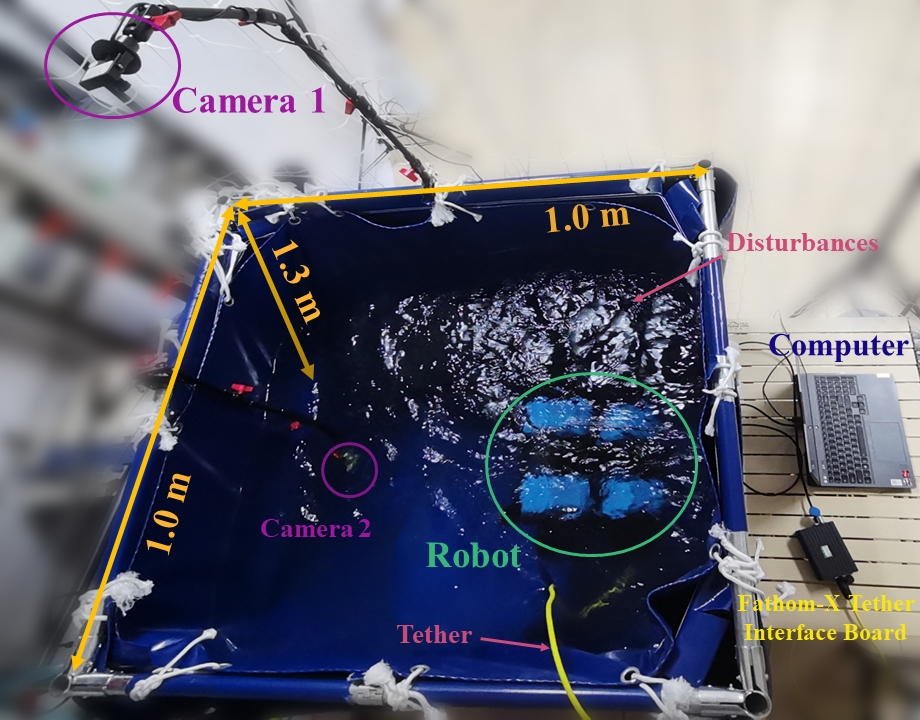}
\caption{A top view snapshot of the experimental setup. The water tank measures 1\,m\,$\times$\,1\,m$\times$\,1.3\,m. Camera \#1 perceives the horizontal positions of the robot, for real-time feedback control. Camera \#2 is employed for recording the experiment from the underwater perspective only. Fathom-X Tether Interface Board facilitates the communication between the robot and upper computer via tethered connections. The confined space provides a testing environment of significant flow disturbances during robot attitude adjustments.}
\label{fig2}
\end{figure}

To create a confined testing underwater environment, the experiment utilizes a water tank measuring 1\,m\,$\times$\,1\,m$\times$\,1.3\,m (length\,$\times$\,width\,$\times$\,height) with a water depth of 1\,m. Figure~\ref{fig2} shows the top view of the experimental water tank. During the experiment, we observed that the wake flows generated by the robot propellers within such limited space resulted in significantly complex turbulent flows. Furthermore, due to the strong coupling between the six degrees of freedom motions when the robot assumes significant pitch angles, simultaneous control of the robot positions in the horizontal plane becomes necessary. To achieve this, we installed a camera (as listed in Table~\ref{tab1}) directly above the tank to obtain real-time horizontal positions of the robot through a color recognition algorithm.

\begin{table}[tb]
\setlength{\tabcolsep}{4.5pt}
\renewcommand{\arraystretch}{1.2}
\scriptsize
\caption{Sensor components used in the experiments and their specifications.}
\begin{center}
\begin{tabular}{@{}cccc@{}}
\toprule
{\footnotesize \textbf{Sensor} }& {\footnotesize\textbf{Model} } & {\footnotesize\textbf{Measured State}}& {\footnotesize\textbf{Sampling Rate}}\\ \midrule
Inertial Measurement Unit & ICM-20689            & Attitude       & 20\,Hz           \\
Pressure Sensor & MS5837-30BA          & Depth       & 10\,Hz           \\
Camera          & G200         & Horizontal Position       & 30\,Hz          \\ \bottomrule
\end{tabular}
\label{tab1}
\end{center}
\end{table}

\section{Dynamic Modeling}\label{section3}

\subsection{Robot Dynamics Model}
\begin{figure}[bp]
\centerline{\includegraphics[width=\linewidth]{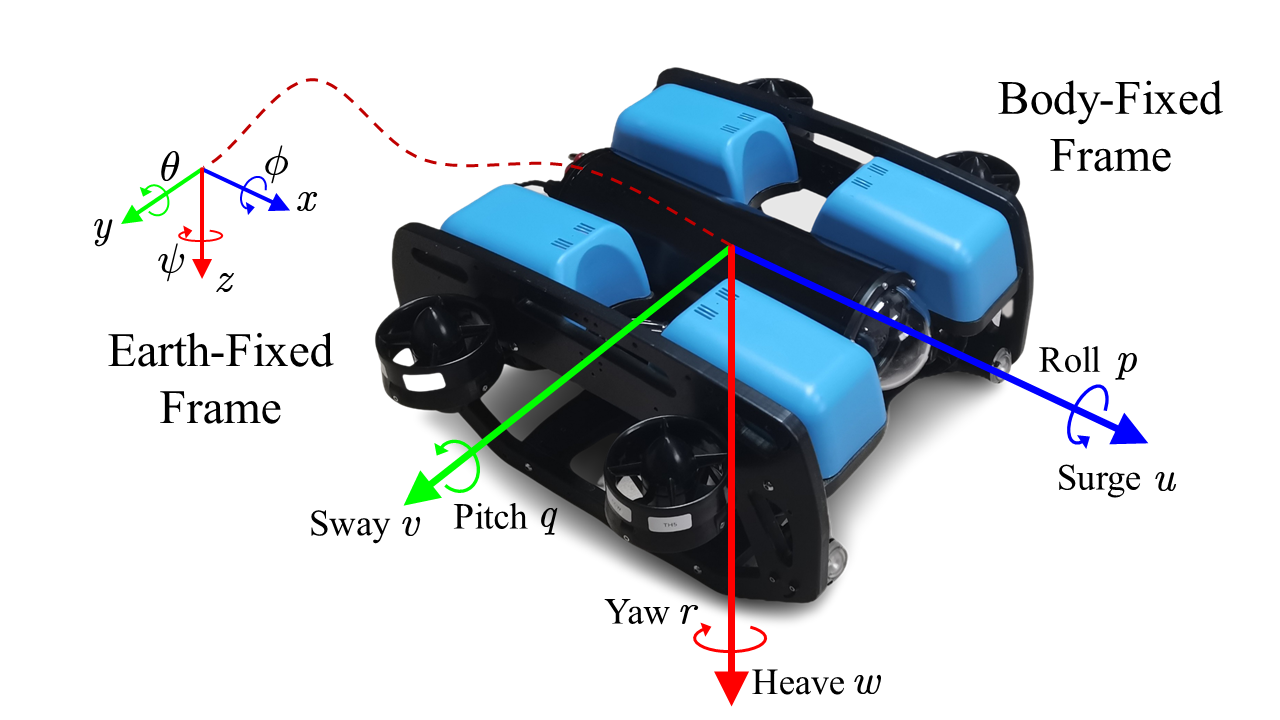}}
\caption{Illustration of the BlueROV2 Heavy, the robot motion states, and the relevant coordinate reference frames.}
\label{fig3}
\end{figure}

This letter adopts the renowned Fossen equation \cite{fossen} to model the dynamics of the underwater robot of interest, following the reference frame definitions and motion variable notations therein, as illustrated in Fig.~\ref{fig3}.

The kinematic transformation from the body-fixed frame to the earth-fixed frame is given by
\begin{equation}
\boldsymbol{\dot{\eta}}=\boldsymbol{J}\left( \boldsymbol{\eta } \right) \boldsymbol{\nu } 
\label{Eq1}
\end{equation} 
where $\boldsymbol{\eta }=\left[ x,y,z,\phi ,\theta ,\psi \right] ^T$ represents the positions and Euler angles expressed in the earth frame, while $\boldsymbol{\nu }=\left[ u,v,w,p,q,r \right] ^T$ represents the velocity of the robot with respect to the earth-fixed frame expressed in the body frame with the six velocity variables corresponding to surge, sway, heave, roll, pitch and yaw. $\boldsymbol{J}\in \mathbb{R}^{6\times 6}$ is the transformation matrix. 

Define $\boldsymbol{\eta }_2=\left[ \phi ,\theta ,\psi \right] ^T$, then we have
\begin{equation}
    \boldsymbol{J}\left( \boldsymbol{\eta } \right) =\left[ \begin{matrix}
	\boldsymbol{J}_1\left( \boldsymbol{\eta }_2 \right)&		\mathbf{0}_{3\times 3}\\
	\mathbf{0}_{3\times 3}&		\boldsymbol{J}_2\left( \boldsymbol{\eta }_2 \right)\\
\end{matrix} \right] 
\label{Eq2}
\end{equation}
Here,
\begin{equation}
\boldsymbol{J}_1\left( \boldsymbol{\eta }_2 \right) =\left[ \begin{matrix}
	c\psi c\theta&		-s\psi c\phi +c\psi s\theta s\phi&		s\psi c\phi +c\psi s\phi s\theta\\
	s\psi c\theta&		c\psi c\phi +s\phi s\theta s\psi&		-c\psi s\phi +s\theta s\psi c\phi\\
	-s\theta&		c\theta s\phi&		c\theta c\phi\\
\end{matrix} \right]
\label{Eq3}
\end{equation}
\begin{equation}
\boldsymbol{J}_2\left( \boldsymbol{\eta }_2 \right) =\left[ \begin{matrix}
	1&		s\phi t\theta&		c\phi t\theta\\
	0&		c\phi&		-s\phi\\
	0&		s\phi /c\theta&		c\phi /c\theta\\
\end{matrix} \right] 
\label{Eq4}
\end{equation}
 where $s$, $c$ and $t$ are abbreviations for sine, cosine and tangent functions, respectively.

The dynamics of the robot is given by 
\begin{equation}
    \boldsymbol{M\dot{\nu}}+\boldsymbol{C}\left( \boldsymbol{\nu } \right) \boldsymbol{\nu }+\boldsymbol{D}\left( \boldsymbol{\nu } \right) \boldsymbol{\nu }+\boldsymbol{g}\left( \boldsymbol{\eta } \right) =\boldsymbol{\tau }+\boldsymbol{\tau }_{\boldsymbol{E}}
\label{Eq5}
\end{equation}
where $\boldsymbol{M}\in\mathbb{R}^{6\times 6}$ represents the inertial matrix, $\boldsymbol{C}\left( \boldsymbol{\nu } \right)\in\mathbb{R}^{6\times 6}$ encompasses centrifugal forces and Coriolis forces, $\boldsymbol{D}\left( \boldsymbol{\nu } \right)\in\mathbb{R}^{6\times 6}$ denotes the damping matrix, $\boldsymbol{g}\left( \boldsymbol{\eta } \right) \in \mathbb{R}^6$ denotes the hydrostatic term, which involves gravity and buoyancy, $\boldsymbol{\tau }\in \mathbb{R}^6$ represents the thruster forces and moments, and $\boldsymbol{\tau }_{\boldsymbol{E}}\in \mathbb{R}^6$ represents the environmental disturbances, specifically, referring to the turbulent flow disturbances occurring within confined space.

It is worth mentioning that the inertial matrix $\boldsymbol{M}$ is of great significance to the controller design. We decompose it into $\boldsymbol{M}=\boldsymbol{M}_{RB}+\boldsymbol{M}_A$, where $\boldsymbol{M}_{RB}$ represents the rigid body inertia matrix, i.e., $\boldsymbol{M}_{RB}=\text{diag}\left( m,m,m,I_x,I_y,I_z \right)$, with $m$ denoting the mass of the robot and $I_x,I_y,I_z$ representing the moments of inertia. $\boldsymbol{M}_A$ refers to the added mass matrix. Assuming a symmetric robot design, we follow the literature \cite{erov} and consider the added mass matrix to be decoupled, i.e.,
$\boldsymbol{M}_A=\text{diag}\left( X_{\dot{u}},Y_{\dot{v}},Z_{\dot{w}},K_{\dot{p}},M_{\dot{q}},N_{\dot{r}} \right)$, where $X_{\dot{u}}\sim N_{\dot{r}}$ represents the added mass and added inertia in the six directions of motion, respectively. This letter adopts the parameter values identified for BlueROV2 Heavy in \cite{phd2015}, as listed in Table~\ref{tab2}.

\subsection{Thruster Model}
The BlueROV2 Heavy is equipped with eight T200 thrusters responsible for controlling all six degrees of freedom motion. Denote the voltage inputs for the thrusters as $\boldsymbol{\mu }=\left[ \mu _1,\mu _2,\mu _3,\mu _4,\mu _5,\mu _6,\mu _7,\mu _8 \right] ^T$ where $\mu _i$ represents the corresponding voltage value for each unit. For ease of discussion, the input voltages are normalized such that $-1\le \mu _i\le 1$. 

The relationship between the propulsion forces and moments $\boldsymbol{\tau }$ provided by the thrusters and their input voltage values $\boldsymbol{\mu }$ is described by
\begin{equation}
    \boldsymbol{\tau }=f_{\max}\boldsymbol{BG}\left( s \right) \boldsymbol{\mu } 
\label{Eq8}
\end{equation}
Here, $\boldsymbol{B}\in\mathbb{R}^{6\times 8}$ represents the thrust allocation matrix, determined by the configuration of thruster placement \cite{OpenAUV2022}. $f_{\max}$ is the maximum allowable thrust for each thruster. $\boldsymbol{G}\left( s \right) \in \mathbb{R}^{8\times 8}$ represents the transfer function matrix from the input voltages of the thrusters to their corresponding output thrusts. Considering that the thruster dynamics is much faster than that of the underwater robot, we neglect the thruster dynamics in subsequent controller design following the literature \cite{ISMC-ECC2021}.

\begin{table}[tb]
\renewcommand{\arraystretch}{1.1}
\caption{Model parameters of the BlueROV2 Heavy.}
\begin{center}
\begin{tabular}{@{}ccc@{}}
\toprule
\textbf{Parameter} & \textbf{Nomenclature} & \textbf{Value} \\ \midrule
Mass               & $m$                     & 13.5\,kg         \\
Moment of Inertia in $x$  &  $I_x$                     & 0.26\,$\text{kgm}^2$\\
Moment of Inertia in $y$         &    $I_y$                    &0.23\,$\text{kgm}^2$\\
Moment of Inertia in $z$          &     $I_z$                   &0.37\,$\text{kgm}^2$\\
Added Mass in $x$    &       $X_{\dot{u}}$                &  6.36\,kg          \\
         Added Mass in $y$          &    $Y_{\dot{v}}$            &  7.12\,kg              \\
         Added Mass in $z$         &    $Z_{\dot{w}}$           &  18.68\,kg             \\
         Added Inertia in $\phi$          &    $K_{\dot{p}}$            &  0.189\,$\text{kgm}^2$              \\
         Added Inertia in $\theta$          &    $M_{\dot{q}}$             &  0.135\,$\text{kgm}^2$              \\
         Added Inertia in $\psi$          &    $N_{\dot{r}}$             &  0.222\,$\text{kgm}^2$              \\
Maximum Thrust of Each Thruster      &   $f_{\max}$            &   15.4\,N           \\ \bottomrule
\end{tabular}
\label{tab2}
\end{center}
\end{table}

\subsection{Dynamic Model with System Uncertainty and Unknown Disturbances}
For the intuition and convenience for the control design, we combine Eqs.~\eqref{Eq1} and~\eqref{Eq5} and obtain the robot dynamics model in the earth-fixed frame, i.e.,
\begin{equation}
    \boldsymbol{M}_{\boldsymbol{\eta }}\boldsymbol{\ddot{\eta}}+\boldsymbol{C}_{\boldsymbol{\eta }}\left( \boldsymbol{\nu ,\eta } \right) \boldsymbol{\dot{\eta}}+\boldsymbol{D}_{\boldsymbol{\eta }}\left( \boldsymbol{\nu ,\eta } \right) \boldsymbol{\dot{\eta}}+\boldsymbol{g}_{\boldsymbol{\eta }}\left( \boldsymbol{\eta } \right) =\boldsymbol{\tau }_{\boldsymbol{\eta }}+\boldsymbol{\tau }_{\boldsymbol{E\eta }}
\label{Eq9}
\end{equation}
where $\boldsymbol{M}_{\boldsymbol{\eta }}=\boldsymbol{J}^{-T}\boldsymbol{MJ}^{-1}
$, $\boldsymbol{D}_{\boldsymbol{\eta }}=\boldsymbol{J}^{-T}\boldsymbol{DJ}^{-1}$, $\boldsymbol{C}_{\boldsymbol{\eta }}=\boldsymbol{J}^{-T}\left( \boldsymbol{C}-\boldsymbol{MJ}^{-1}\boldsymbol{\dot{J}} \right) \boldsymbol{J}^{-1}$,  $\boldsymbol{g}_{\boldsymbol{\eta }}\left( \boldsymbol{\eta } \right) =\boldsymbol{J}^{-T}\boldsymbol{g}\left( \boldsymbol{\eta } \right) 
$, $\boldsymbol{\tau }_{\boldsymbol{\eta }}=\boldsymbol{J}^{-T}\boldsymbol{\tau }
$ and $\boldsymbol{\tau }_{\boldsymbol{E\eta }}=\boldsymbol{J}^{-T}\boldsymbol{\tau }_{\boldsymbol{E}}
$.

We rewrite the dynamics model (Eq.~\eqref{Eq9}) in a compact form, i.e.,
\begin{equation}
    \boldsymbol{\ddot{\eta}}=\boldsymbol{f}\left( \boldsymbol{\eta ,\dot{\eta}} \right) +\boldsymbol{\tilde{\tau}}+\boldsymbol{\tilde{\tau}}_{\boldsymbol{E}}
\label{Eq10}
\end{equation}
where $\boldsymbol{f}\left( \boldsymbol{\eta ,\dot{\eta}} \right)$ represents the nominal inherent dynamics of the system. $\boldsymbol{\tilde{\tau}}$ denotes the control input calculated as
\begin{equation}
    \boldsymbol{\tilde{\tau}}=\left( \boldsymbol{J}^{-T}\left( \boldsymbol{\eta } \right) \boldsymbol{MJ}^{-T}\left( \boldsymbol{\eta } \right) \right) ^{-1}\boldsymbol{J}^{-T}\left( \boldsymbol{\eta } \right) \boldsymbol{\tau }
\label{Eq11}
\end{equation}
$\boldsymbol{\tilde{\tau}}_{\boldsymbol{E}}$ represents the flow disturbances given by
\begin{equation}
    \boldsymbol{\tilde{\tau}}_{\boldsymbol{E}}=\left( \boldsymbol{J}^{-T}\left( \boldsymbol{\eta } \right) \boldsymbol{MJ}^{-T}\left( \boldsymbol{\eta } \right) \right) ^{-1}\boldsymbol{J}^{-T}\left( \boldsymbol{\eta } \right) \boldsymbol{\tau }_{\boldsymbol{E}}
\label{Eq12}
\end{equation}

Taking into account the uncertainty in the system dynamics model, Eq.~\eqref{Eq10} is rewritten as
\begin{equation}
\begin{aligned}
    \boldsymbol{\ddot{\eta}}&=\boldsymbol{f}\left( \boldsymbol{\eta ,\dot{\eta}} \right) +\Delta \boldsymbol{f}\left( \boldsymbol{\eta ,\dot{\eta}} \right) +\boldsymbol{\tilde{\tau}}+\boldsymbol{H}_1\left( \boldsymbol{\eta ,\dot{\eta}} \right) \boldsymbol{\tilde{\tau}}\\&+\boldsymbol{\tilde{\tau}}_{\boldsymbol{E}}+\boldsymbol{H}_2\left( \boldsymbol{\eta ,\dot{\eta}} \right) \boldsymbol{\tilde{\tau}}_{\boldsymbol{E}}
\end{aligned}
\label{Eq13}
\end{equation}
where $\Delta \boldsymbol{f}$,  $\boldsymbol{H}_1\in \mathbb{R}^{6\times 6}$, and $\boldsymbol{H}_2\in \mathbb{R}^{6\times 6}$ represent the uncertainties in $\boldsymbol{f}$, $\boldsymbol{\tilde{\tau}}$ and $\boldsymbol{\tilde{\tau}}_{\boldsymbol{E}}$, respectively.

Grouping relevant terms together, we obtain the following dynamics model
\begin{equation}
    \boldsymbol{\ddot{\eta}}=\boldsymbol{\tilde{\tau}}+\boldsymbol{d}\left( \boldsymbol{\eta ,\dot{\eta},}t \right) 
\label{Eq14}
\end{equation}
where $\boldsymbol{d}\left( \boldsymbol{\eta ,\dot{\eta},}t \right)$ encompasses all terms on the right-hand side of Eq.~\eqref{Eq13} excluding $\boldsymbol{\tilde{\tau}}$.

\section{AISMC Controller}\label{section4}

\subsection{ISMC Module}

Controlling the attitudes of underwater robots over a wide range of pitch angles presents challenges, potentially resulting in significant steady-state errors due to time-varying unstructured flow disturbances and non-equilibrium regulation states. To address these issues, this section introduces the ISMC module in AISMC, as depicted in Fig.~\ref{fig1}. It is noteworthy that while our goal is attitude tracking control for a station-holding underwater robot, the coupled dynamics across all directions of motion, especially at large pitch angles, necessitate simultaneous control of all six degrees of freedom. We define the reference trajectory as $\boldsymbol{\eta _r}=\left[ x_r,y_r,z_r,\phi _r,\theta _r,\psi _r \right] ^T$. The tracking error $\boldsymbol{e}\in \mathbb{R}^6$ is given by
 \begin{equation}
    \boldsymbol{e}=\boldsymbol{\eta }-\boldsymbol{\eta }_{\text{r}}
\label{Eq15}
\end{equation}

The sliding surface $\boldsymbol{s}\in \mathbb{R}^6$ is then designed as
\begin{equation}
\boldsymbol{s}=\boldsymbol{\dot{e}}+\boldsymbol{C}_1\boldsymbol{e}+\boldsymbol{C}_2\int{\boldsymbol{e}\text{d}t}
\label{Eq16}
\end{equation}
where $\boldsymbol{C}_i=\text{diag}\left( c_{i,x},c_{i,y},c_{i,z},c_{i,\phi},c_{i,\theta},c_{\psi} \right) $, $i=1,2$ denotes the proportional and integral gains, respectively. The objective of ISMC is to ensure that $\boldsymbol{s}$ reaches and remains on the sliding surface $\boldsymbol{s}=0$ in finite time. Given the elements of $\boldsymbol{C}_1$ and $\boldsymbol{C}_2$ are positive, it is straightforward to show by Eq.~\eqref{Eq16} that both $\boldsymbol{e}$ and $\boldsymbol{\dot{e}}$ exponentially converge to $\mathbf{0}$.

The control law is designed as
\begin{equation}
    \boldsymbol{\tilde{\tau}}=\boldsymbol{\ddot{\eta}}_r-\boldsymbol{C}_1\boldsymbol{\dot{e}}-\boldsymbol{C}_2\boldsymbol{e}-\boldsymbol{\varGamma s}-\boldsymbol{K}\text{sgn} \left( \boldsymbol{s} \right) 
\label{Eq17}
\end{equation}
where the sign function $\text{sgn} \left( \boldsymbol{s} \right)$ represents the switching behavior, $\boldsymbol{K}=\text{diag}\left( k_x,k_y,k_z,k_{\phi},k_{\theta},k_{\psi} \right)$ represents the switching gain. The gain $\boldsymbol{\varGamma }=\text{diag}\left( \gamma _x,\gamma _y,\gamma _z,\gamma _{\phi},\gamma _{\theta},\gamma _{\psi} \right)$ is introduced to expedite the exponential convergence.

\textbf{\textit{Stability Analysis:}}
This letter uses Lyapunov analysis to establish the stability of the ISMC closed-loop control system. Define the Lyapunov function as
\begin{equation}
    V_1=\frac{1}{2}\boldsymbol{s}^T\boldsymbol{s}
\label{Eq18}
\end{equation}

Combined with Eqs.~\eqref{Eq14}--\eqref{Eq16}, the time derivative of $V_1$ is calculated as
\begin{equation}
    \begin{aligned}
        \dot{V}_1&=\boldsymbol{s}^T\boldsymbol{\dot{s}} =\boldsymbol{s}^T\left( \boldsymbol{\tilde{\tau}}+\boldsymbol{d}\left( \boldsymbol{\eta ,\dot{\eta},}t \right) -\boldsymbol{\ddot{\eta}}_r+\boldsymbol{C}_1\boldsymbol{\dot{e}}+\boldsymbol{C}_2\boldsymbol{e} \right) \\
    \end{aligned}
\label{Eq19}
\end{equation}

Substituting \eqref{Eq17} into \eqref{Eq19} yields
\begin{equation}
    \begin{aligned}
        \dot{V}_1&=\boldsymbol{s}^T\left( -\boldsymbol{\varGamma s}-\boldsymbol{K}\text{sgn} \left( \boldsymbol{s} \right) +\boldsymbol{d}\left( \boldsymbol{\eta ,\dot{\eta},}t \right) \right) \\
               &=-\boldsymbol{s}^T\boldsymbol{\varGamma s}-\sum_{i\in l}{\left( k_i-d_i\left( \boldsymbol{\eta ,\dot{\eta},}t \right) \right)}\left| s_i \right|
    \end{aligned}
\label{Eq20}
\end{equation}

If there exists $k_i\ge d_i$ for all $i\in l=\left\{ x,y,z,\phi ,\theta ,\psi \right\}$, then
\begin{equation}
    \dot{V}_1\le -\boldsymbol{s}^T\boldsymbol{\varGamma s}\le 0
\label{Eq21}
\end{equation}

The condition $V_1=0$ is satisfied only when $\boldsymbol{s}=\mathbf{0}$. In accordance with LaSalle's invariance principle, the closed-loop system is asymptotically stable, and there exists a finite time $t_f$ such that $\boldsymbol{s}\rightarrow \mathbf{0}$ as $t\rightarrow t_f<\infty$. 

\subsection{AISMC Design}
The stability analysis above reveals that for the ISMC system to maintain stable, it is imperative to ensure that the switching gains surpass the upper bound of the unknown system dynamics. However, determining this upper bound beforehand is extremely challenging in practical scenarios. Both system dynamics and external disturbances exhibit time-varying characteristics. Consequently, adopting high switching gains as a conservative control strategy for stability concerns often results in degraded control performance and severe chattering, compromising control accuracy and potentially causing propeller failure. To address these issues, we propose the AISMC control strategy, introducing adaptability into the switching gain. The control law is given by
\begin{equation}
        \boldsymbol{\tilde{\tau}}=\boldsymbol{\ddot{\eta}}_r-\boldsymbol{C}_1\boldsymbol{\dot{e}}-\boldsymbol{C}_2\boldsymbol{e}-\boldsymbol{\varGamma s}-\boldsymbol{\hat{K}}\text{sgn} \left( \boldsymbol{s} \right) 
\label{Eq22}
\end{equation}
where $\boldsymbol{\hat{K}}=\text{diag}\left(\hat{k}_x,\hat{k}_y,\hat{k}_z,\hat{k}_{\phi},\hat{k}_{\theta},\hat{k}_{\psi} \right)$. The adaptive law is designed as
\begin{equation}
    \dot{\hat{k}}_i=\begin{cases}
    \bar{k}_i\left| s_i \right|\cdot \text{sgn} \left( \left| s_i-\epsilon _i \right| \right) \ &\text{if\ }\hat{k}_i>\beta _i\\
	\beta _i\ &\text{if\ }\hat{k}_i\le \beta _i\\
\end{cases}
\label{Eq23}
\end{equation}
where $\bar{k}_i>0$ represents the changing rate of adaptive gain $\hat{k}_i$. $\beta _i$, a very small positive value, is set to ensure that $\hat{k}_i$ remains positive. The boundary layer, denoted as $\epsilon _i$, is designed considering the discrete-time control implemented on the real underwater robot. In this scenario, $s_i$ cannot be strictly equal to 0; instead, it is considered that the sliding surface is attained when $\left| s_i \right|\le \epsilon _i$.

When the system states deviate significantly from the sliding surface, the gain $\hat{k}_i$ rapidly increases to promptly drive the states back to the sliding surface. Once the sliding surface is attained, $\hat{k}_i$ decreases to mitigate the chattering problem. Subsequently, if $\hat{k}_i$ becomes insufficient to counteract external disturbances, causing a tendency for these disturbances to pull the system away from the sliding surface, $\hat{k}_i$ increases again to effectively combat such disturbances. In other words, $\hat{k}_i$ is maintained at a minimum level that allows for establishment of the sliding pattern and precisely resistance to uncertain disturbances.

\textbf{\textit{Stability Analysis:}}
The stability of the AISMC system is analyzed by defining a new Lyapunov function, i.e.,
\begin{equation}
    V_2=\frac{1}{2}\boldsymbol{s}^T\boldsymbol{s}+\sum_{i\in l}{\frac{1}{2\alpha _i}\left( \hat{k}_i-\sup\ \hat{k}_i \right) ^2}
\label{Eq24}
\end{equation}
where $\alpha_i$ is a positive constant, and there exists a finite supremum of $\hat{k}_i$ \cite{ASMC2013}. Following the Lyapunov analysis \cite{ASMC2010}, it is straightforward to show the ultimate boundedness of the tracking error. Specifically, within a finite time, $s_i$ converges towards a vicinity $\delta_i$, slightly larger than $\epsilon_i$, taking the form
\begin{equation}
    \left| s_i \right|\le \delta _i=\sqrt{\epsilon _{i}^{2}+\frac{\xi _i}{\bar{k}_i}}
\label{Eq25}
\end{equation}
where $\xi _i$ is associated with the upper bound of robot dynamics and disturbances.

In this design, the selection of the value of $\epsilon_i$ is crucial. A poor choice may lead to system divergence. It is generally preferable to opt for a larger value rather than a smaller one. We design $\epsilon_i$ to be proportional to $\hat{k}_i$, i.e.,
\begin{equation}
    \hat{\epsilon}_i=\lambda _i\hat{k}_i
\label{Eq26}
\end{equation}
where the positive number $\lambda_i$ is a tunable design parameter.

The AISMC control law (Eqs.~\eqref{Eq22},~\eqref{Eq23}, and~\eqref{Eq26}), provides a comprehensive solution to the attitude control of station-holding underwater robots in confined space. Through Eqs.~\eqref{Eq8} and~\eqref{Eq11}, the input voltage to each thruster is ultimately determined.

\section{Experimental Validation and Analysis}\label{section5}
The proposed AISMC control algorithm is experimentally validated using BlueROV2 Heavy and the experimental setup described in Section~\ref{section2}. This section introduces the control tasks and proceeds to discuss and analyze the experimental results.

\subsection{Control Tasks}\label{section5-1}
We design three different control tasks to validate the proposed AISMC algorithm. Specifically, these tasks involve the robot following reference trajectories of attitude angles, in particular, the pitch angle, including zero-angle holding, step trajectory following, and sine trajectory tracking. In all the control tasks, the robot is required to maintain a fixed position, i.e., $\dot{x}_r=0,\ \dot{y}_r=0\ \text{and\ }\dot{z}_r=0$.

\textbf{Task \#1} Zero-angle holding: Maintaining zero angles is fundamental for attitude control of underwater robots. The reference trajectory of attitude is defined as $\theta _r=0,\ \phi _r=0\ \text{and\ }\psi _r=0$.

\textbf{Task \#2} Step trajectory following: In confined space, robots must possess the capability to adapt their attitudes in accordance with environmental constraints. This task selects a step function $\theta _r=\frac{\pi}{2}$ as the reference trajectory to evaluate the robot's performance in regulating and maintaining a nonzero pitch angle. At the same time, the robot maintains stable roll and yaw, i.e., $\ \phi _r=0\ \text{and\ }\psi _r=0$.

\textbf{Task \#3} Sine trajectory tracking: To evaluate the control performance in tracking a continuously changing pitch angle, this task selects a reference trajectory of a sinusoidal function with a wide range of pitch angle variation. The reference trajectory is defined as follows: $\theta _r=\frac{\pi}{4}\sin \left( \frac{\pi}{120}\left( t-5 \right) \right) $ if $t\in(5,125)$ and $\theta_r=0$ otherwise; $\phi _r=0\ \text{and\ }\psi _r=0$.

\subsection{Results and Analysis}\label{section5-3}
This section presents the experimental results and conducts an in-depth analysis. The specific parameters employed for AISMC in our experiment are detailed in Table~\ref{tab3}. To illustrate the effectiveness of the proposed method, we adopt two other benchmark control schemes for comparative studies, i.e., Proportional-Integral-Derivative (PID) Control and Conventional Sliding Mode Control (SMC) methods. In addition, we include the simulation results of AISMC. To more accurately simulate the disturbance in confined space, the flow disturbance term $\boldsymbol{\tau }_{\boldsymbol{E}}$ in Eq.~\eqref{Eq5} is modeled as Gaussian process. 

\begin{table}[tb]
\renewcommand{\arraystretch}{1.1}
\caption{Control parameters used in the AISMC experiment.}
\begin{center}
\begin{tabularx}{\linewidth}{*{6}{>{\centering\arraybackslash}X}}
\toprule
$i$    &$c_{1,i}$    &$c_{2,i}$  &$\gamma _i$  &$\bar{k}_i$  &$\lambda _i$  \\ \midrule
$x$ &1.4  &2  &0.2  &0.15  &20  \\
$y$ &1.6  &2  &0.2  &0.1 &20  \\
$z$  &2  &0.7  &0.2  &0.015  &20  \\
$\phi$  &0.7  &1  &8  &0.15  &20  \\
$\theta$  &0.85  &0.8  &8  &0.15  &20  \\
$\psi$  &2  &1.5  &8  &0.025  &20  \\ \bottomrule
\end{tabularx}
\label{tab3}
\end{center}
\end{table}

Figure~\ref{fig5} presents the experimental results of attitude trajectories in Task \#1. We observe that AISMC effectively maintains zero attitude angles. Specifically, the tracking errors for pitch and roll consistently remain within 0.02\,rad, whereas for yaw, the error is confined within 0.05\,rad. Importantly, AISMC exhibits significantly reduced control oscillations compared to PID and SMC, particularly in pitch and roll, demonstrating its robust control performance against time-varying uncertain flow disturbances. Additionally, the experimental results match the simulated trajectories with almost negligible deviations, confirming the accuracy of the established dynamics model and the subsequent  control design.

\begin{figure*}[!t]
    \centering
    \subfloat[$t=0\,s$]{\includegraphics[width=0.1925\linewidth]{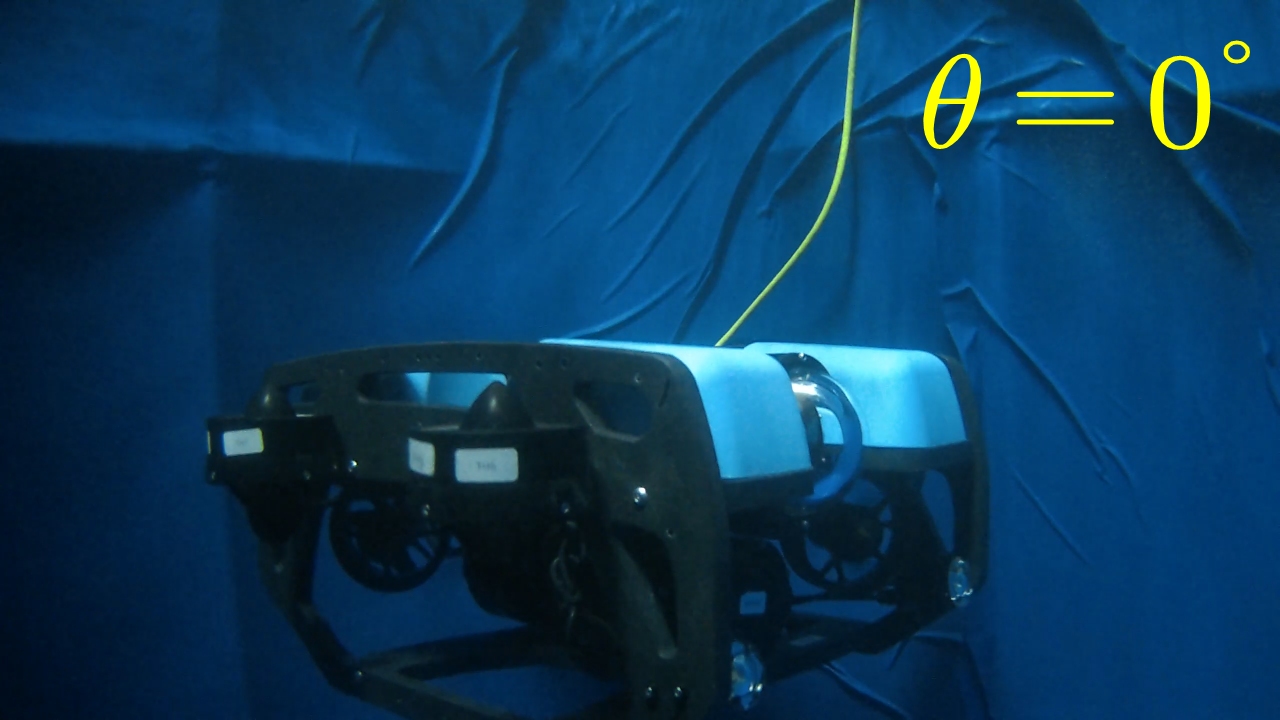}\label{fig:t0}}
    \hfil
    \subfloat[$t=35\,s$]{\includegraphics[width=0.1925\linewidth]{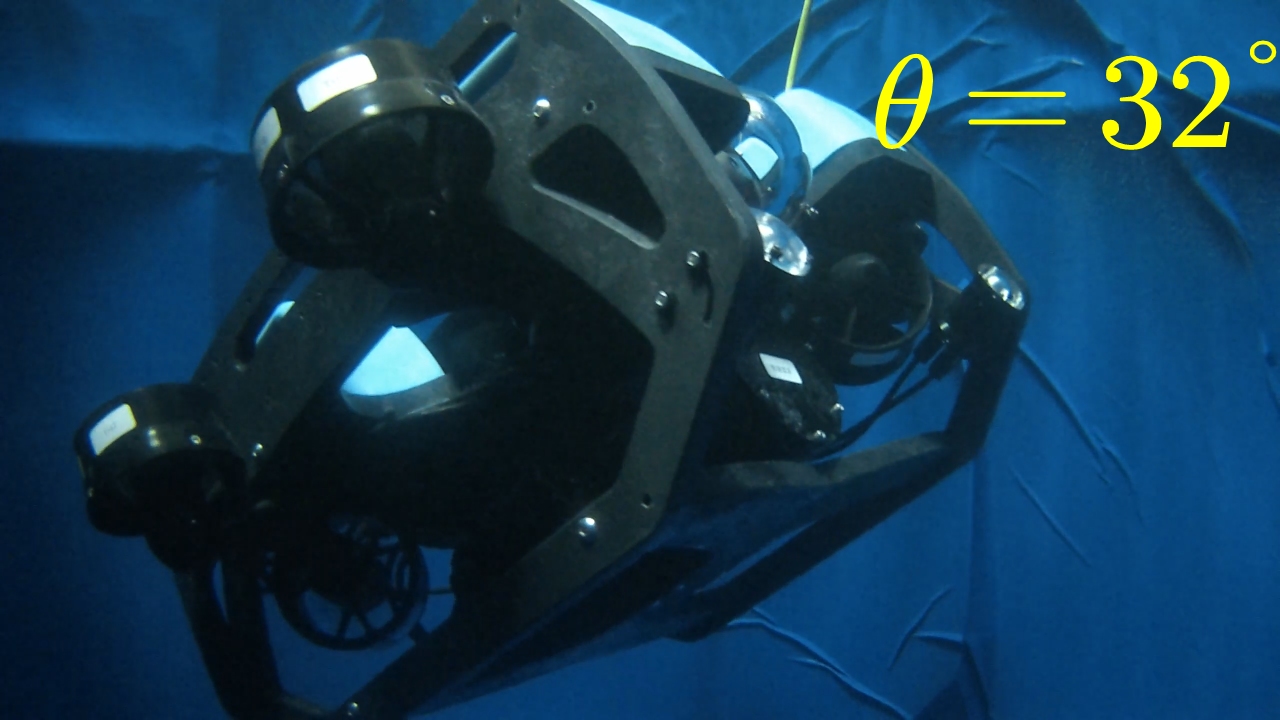}\label{fig:t35}}
    \hfil
    \subfloat[$t=65\,s$]{\includegraphics[width=0.1925\linewidth]{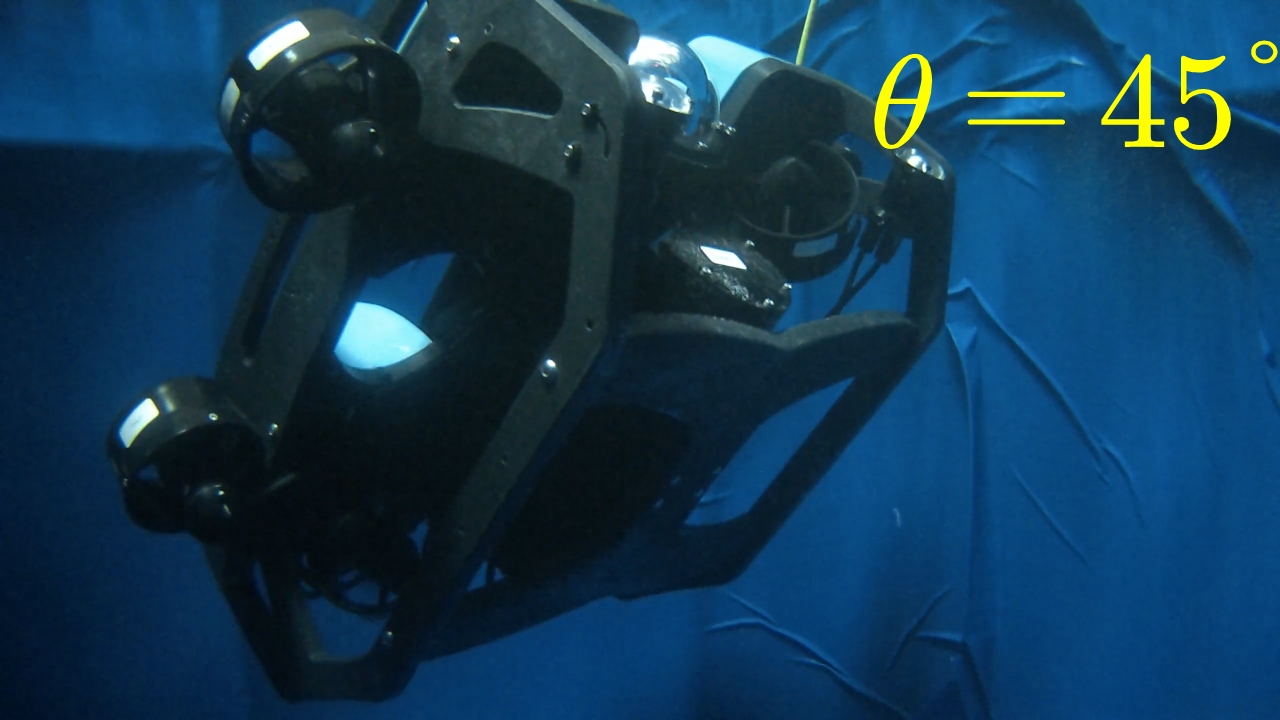}\label{fig:t65}}
    \hfil
    \subfloat[$t=95\,s$]{\includegraphics[width=0.1925\linewidth]{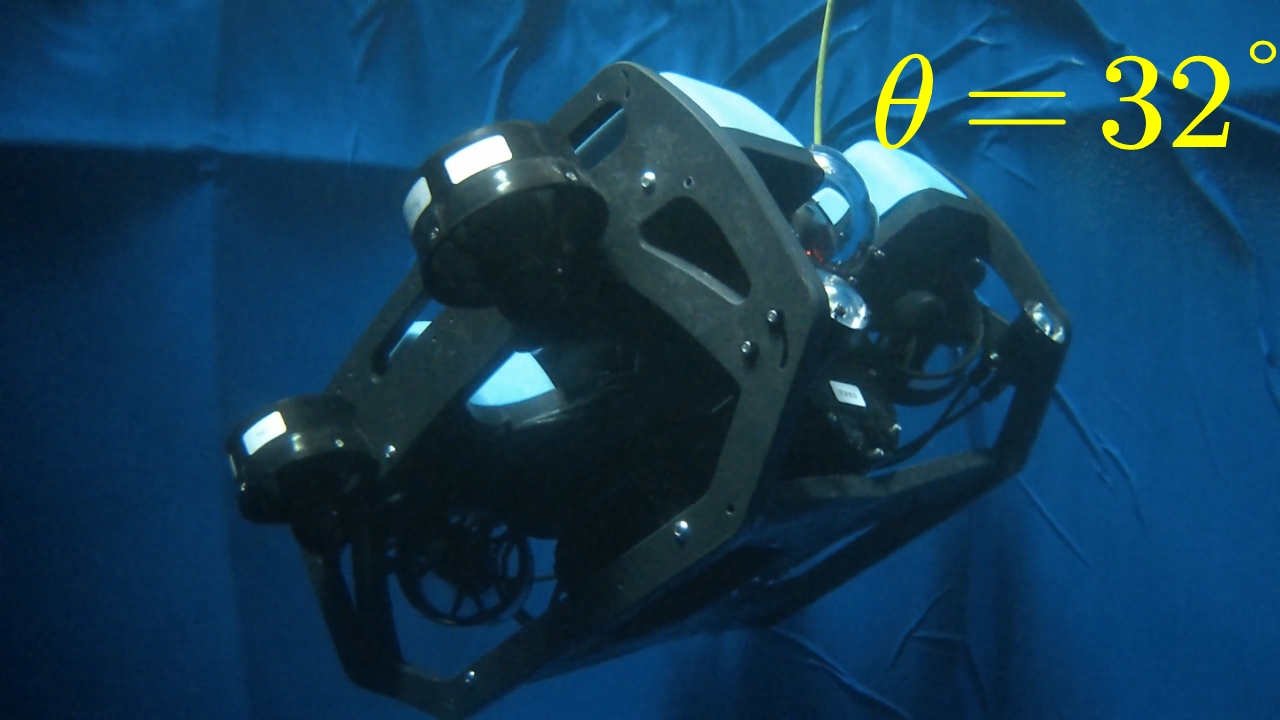}\label{fig:t95}}
    \hfil
    \subfloat[$t=130\,s$]{\includegraphics[width=0.1925\linewidth]{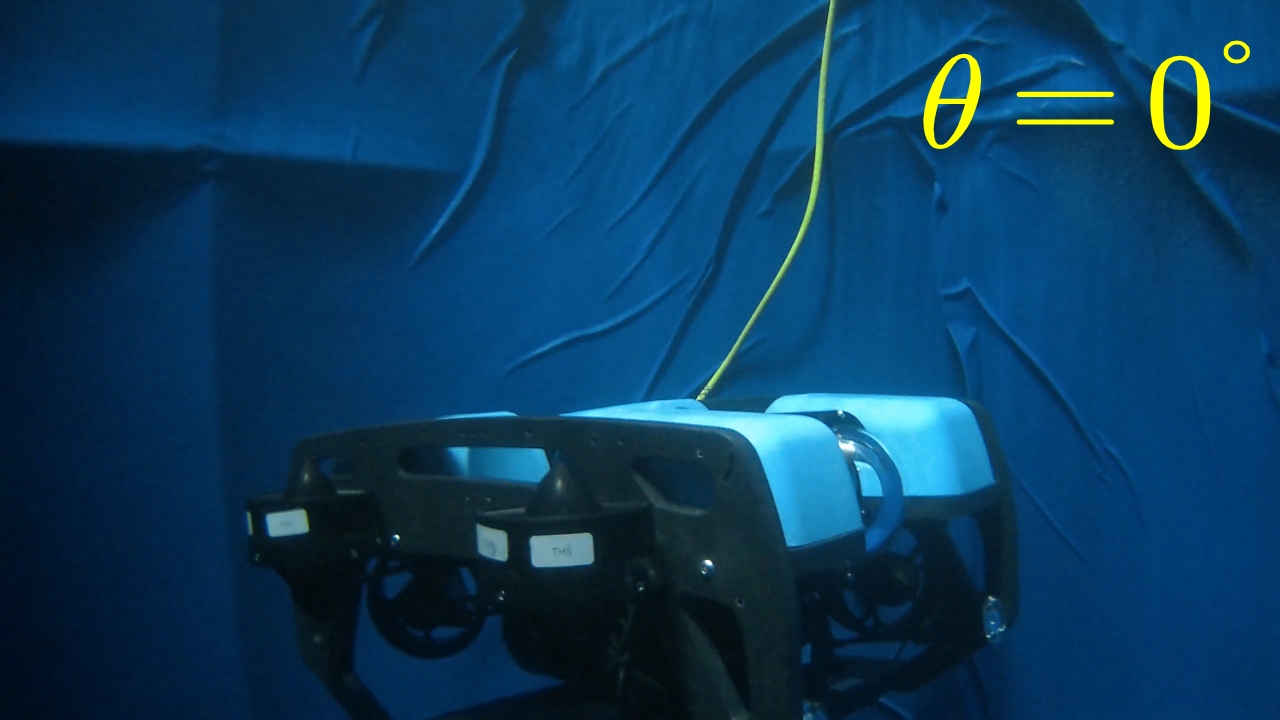}\label{fig:t130}}
    \caption{The robot undergoing large pitch variations controlled by AISMC in Task \#3.}
    \label{fig:evolution}
\label{fig4}
\end{figure*}

\begin{figure}[tbp]
\centerline{\includegraphics{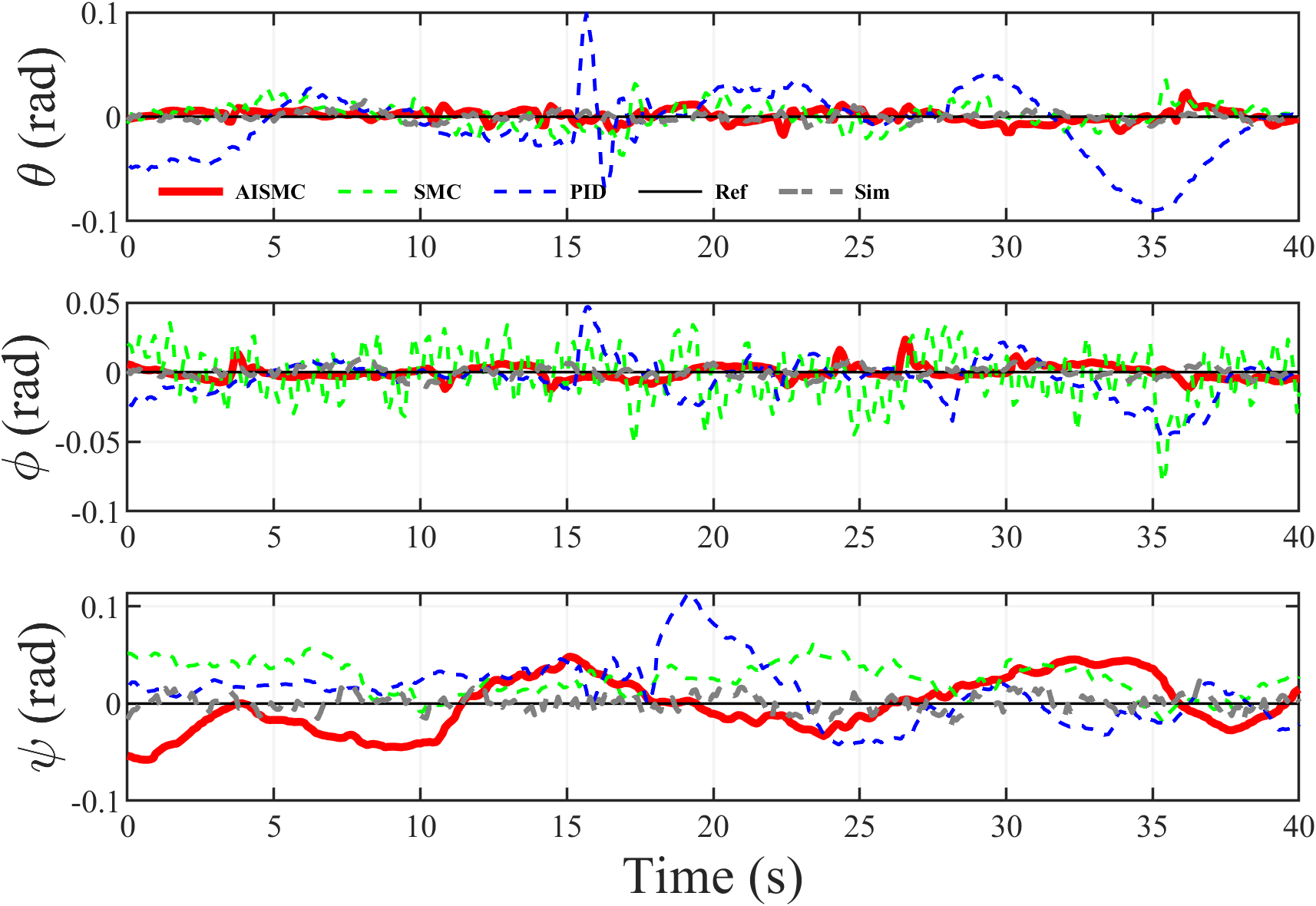}}
\caption{Attitude tracking performance in Task \#1. "Ref" denotes the reference trajectory, "Sim" represents the results of AISMC simulation, and the remaining curves show the experimental results of different controllers. Similarly for Fig.~\ref{fig6} and Fig.~\ref{fig7}.}
\label{fig5}
\end{figure}

\begin{figure}[tbp]
\centerline{\includegraphics{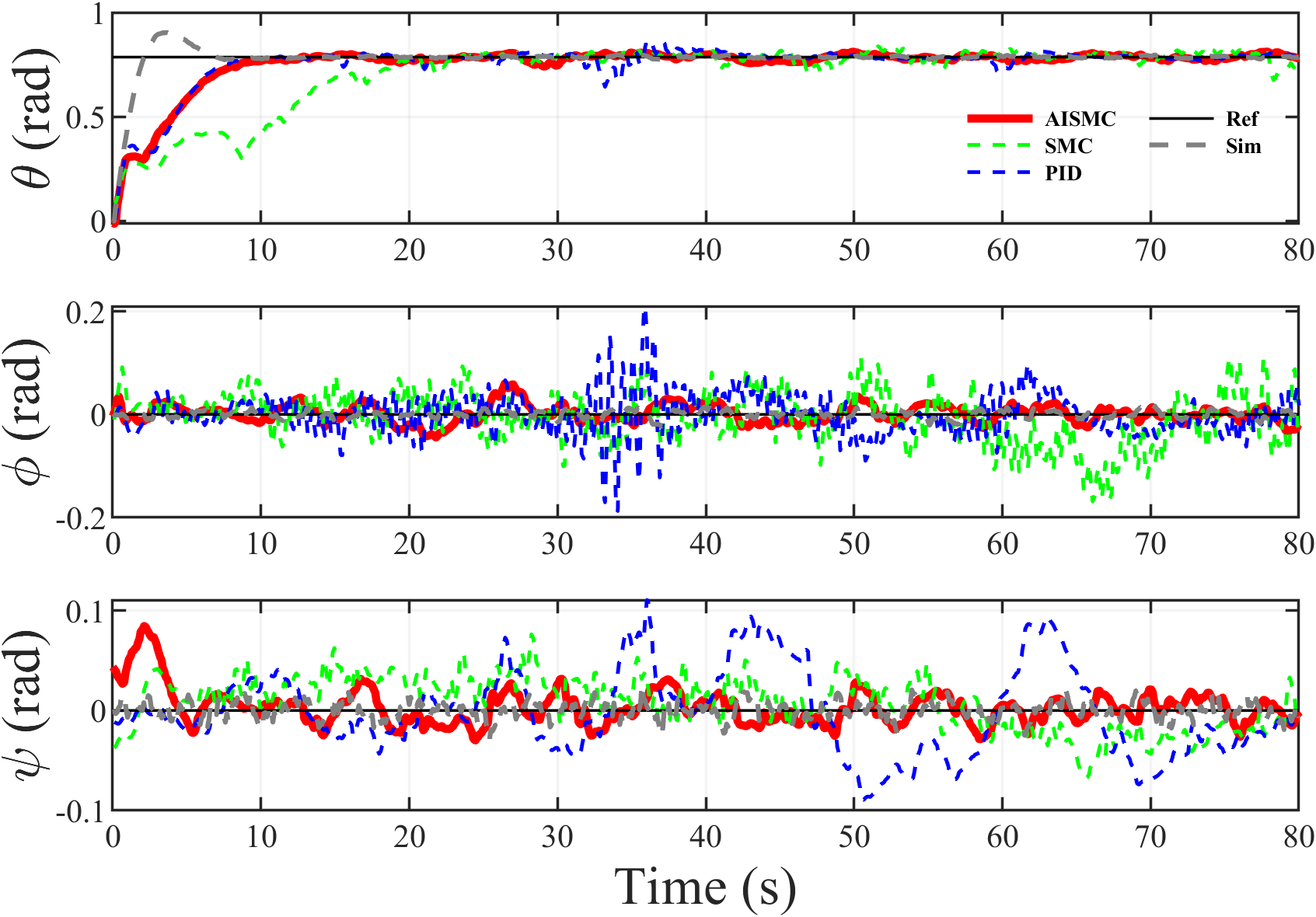}}
\caption{Attitude tracking performance in Task \#2.}
\label{fig6}
\end{figure}

Figure~\ref{fig6} presents the experimental results of attitude trajectories in Task \#2, following a step reference trajectory. It is observed that the AISMC successfully executed a rapid ascent from \ang{0} to \ang{45} in the pitch angle within a rise time of less than 10\,seconds during the experimental process. Moreover, at the sustained \ang{45} pitch angle, AISMC demonstrated excellent stability, with both roll and yaw tracking errors being controlled within a range of 0.03\,rad. In contrast, SMC exhibits a rise time of up to 20\,seconds, and significant chattering during the subsequent \ang{45} holding, with roll amplitude exceeding 0.1\,rad. Although PID demonstrates a quick rise time, significant oscillations in the controlled attitude occur around 30\,s, presumably due to its poor disturbance rejection capability.

\begin{figure}[tbp]
\centerline{\includegraphics{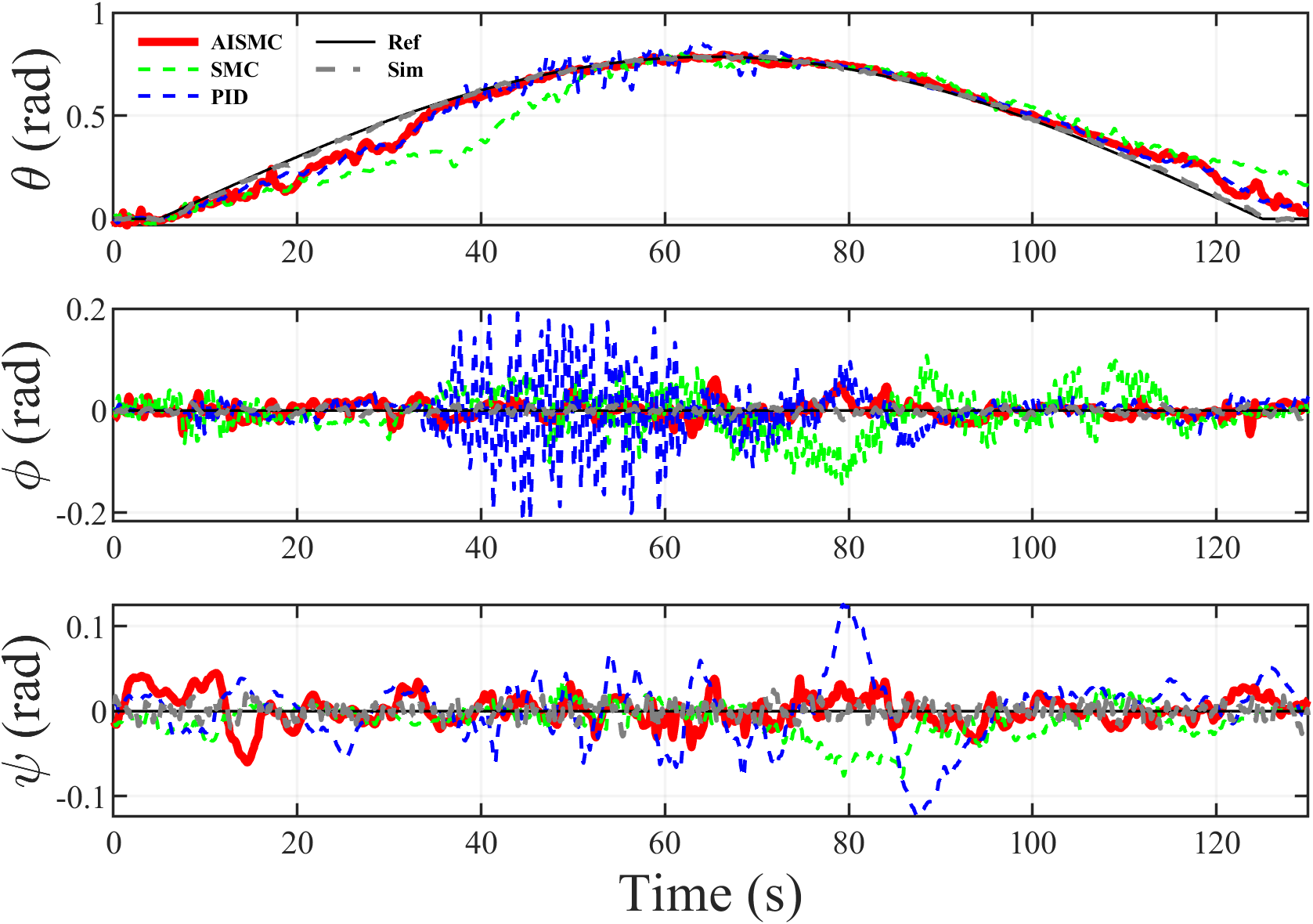}}
\caption{Attitude tracking performance in Task \#3.}
\label{fig7}
\end{figure}

The experimental results of attitude control in Task \#3 are illustrated in Fig.~\ref{fig7}. Following a sinusoidal reference trajectory, SMC exhibits poor tracking accuracy across all the three attitude angles with large tracking errors up to 0.3\,rad. PID, on the other hand, experiences severe oscillations around 40--60\,s, with a roll angle oscillation amplitude reaching 0.2\,rad. Conversely, AISMC demonstrates satisfactory tracking accuracy for the sinusoidal reference, with only minor deviations observed at certain instances such as $t=30\,s$. Moreover, throughout such large-range continuous pitch variations, roll and yaw angles maintain close to the desired values with fluctuations within 0.04\,rad. Figure~\ref{fig4} showcases the AISMC-controlled robot executing a pitch angle change from \ang{0} to \ang{45} and back to \ang{0}, illustrating its ability to manage such variations effectively while maintaining satisfactory tracking accuracy.

\begin{figure}[tbp]
\centerline{\includegraphics[width=0.95\linewidth]{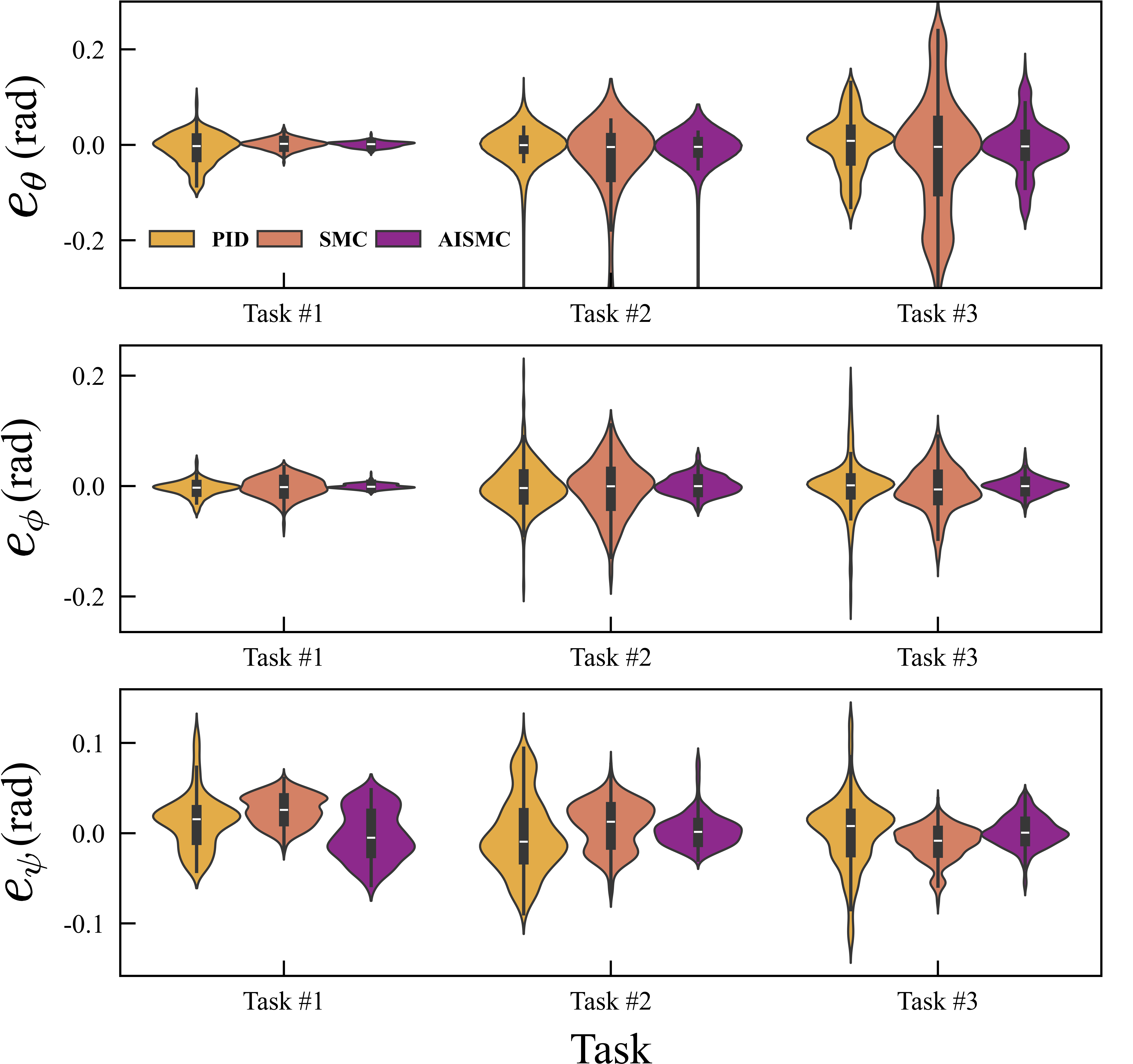}}
\caption{Violin plot comparing the error distributions of three controllers.}
\label{fig8}
\end{figure}

Overall, AISMC demonstrates superior responsiveness, enhanced robustness, and markedly reduced chattering, with its integral component and adaptive aspects playing pivotal roles. Figure~\ref{fig8} presents a violin plot of trajectory tracking errors, providing a more intuitive visualization of the error distribution across different tasks and controllers. As shown, the violin shapes for AISMC tend to be flatter, indicating that the tracking errors are more tightly concentrated around zero, which aligns with the previous observations. Table~\ref{tab4} lists the Root Mean Square Error (RMSE) for all control experiments. It is observed that when compared to PID, AISMC reduces the total RMSE by 41.0\%, 6.3\%, and 29.0\% for the three tasks respectively; while compared to SMC, the total RMSE reduction is 23.0\%, 36.1\%, and 54.4\%. These findings indicate that the proposed AISMC controller successfully achieves stable attitude control during large-range pitch variations in confined space.

\begin{table}[tb]
\renewcommand{\arraystretch}{1.25}
\caption{RMSE for different tasks and controllers. The "total" column represents the RMSE calculated jointly for all three attitude angles.}
\begin{center}
\begin{tabular}{@{}cccccc@{}}
\toprule
\multirow{2}{*}{\textbf{Task}} & \multirow{2}{*}{\textbf{Controller}} & \multicolumn{4}{c}{\textbf{RMSE}} \\ \cmidrule(l){3-6} 
                               &                                      & Pitch  & Roll   & Yaw    & Total  \\ \midrule
\multirow{3}{*}{   Task \#1}            & PID                                  & 0.0330 & 0.0149 & 0.0331 & 0.0283 \\
                               & SMC                                  & 0.0110 & 0.0186 & 0.0307 & 0.0217 \\
                               & AISMC                                & 0.0060 & 0.0049 & 0.0279 & 0.0167 \\ \midrule
\multirow{3}{*}{  Task \#2}            & PID                                  & 0.1144 & 0.0401 & 0.0419 & 0.0741 \\
                               & SMC                                  & 0.1795 & 0.0495 & 0.0275 & 0.1086 \\
                               & AISMC                                & 0.1175 & 0.0175 & 0.0184 & 0.0694 \\ \midrule
\multirow{3}{*}{  Task \#3}            & PID                                  & 0.0582 & 0.0473 & 0.0393 & 0.0489 \\
                               & SMC                                  & 0.1234 & 0.0408 & 0.0221 & 0.0761 \\
                               & AISMC                                & 0.0555 & 0.0153 & 0.0174 & 0.0347 \\ \bottomrule
\end{tabular}
\label{tab4}
\end{center}
\end{table}

It is noteworthy that the enhancement in AISMC yaw control relative to SMC and PID is minimal during Task~\#1 with a significantly higher tracking error than pitch and roll. We speculate two factors contribute to this phenomenon. First, the yaw angle measurement, relying on a compass sensor, is susceptible to external magnetic field interference. Second, yaw control is strongly coupled with vision-based X and Y position control at zero attitude angles, prone to light disturbances affecting position measurement accuracy and yaw control performance.


\section{Conclusion}\label{section6}
This letter proposed an adaptive integral sliding mode control (AISMC) algorithm for attitude control of underwater robots navigating confined space. Different from conventional studies that primarily consider yaw angle control, the proposed method enabled simultaneous control over pitch, roll, and yaw angles with a wide range of pitch variations. This task is highly challenging due to the strong coupling across six degrees of freedom, particularly the presence of unknown and complex turbulence within confined space. Lyapunov analysis was applied to rigorously establish the stability of the closed-loop control system. Extensive experiments and comparison studies were conducted, the results of which validated the effectiveness of the proposed AISMC controller. In addition, AISMC significantly outperformed conventional PID and SMC approaches in terms of tracking accuracy, chattering elimination, and robustness against disturbances.

In future work, we will refine dynamics modeling and turbulence estimation to enhance AISMC control performances. Ultimately, we plan to deploy the AISMC algorithm in real-world aquatic environments, supporting missions such like underwater scientific exploration.

\section*{Acknowledgment}
The authors gratefully acknowledge the support of the Beijing Municipal Natural Science Foundation (No.QY23047).

\bibliographystyle{IEEEtran}
\bibliography{IEEEabrv,paper}

\end{document}